\documentclass[conference]{IEEEtran}
\IEEEoverridecommandlockouts
\renewcommand\IEEEkeywordsname{Keywords}
\usepackage{cite}
\usepackage{amsmath,amssymb,amsfonts}
\usepackage{algorithmic}
\usepackage{graphicx}
\usepackage{textcomp}

\usepackage[table]{xcolor}
\usepackage{hyperref}

\newtheorem{theorem}{Theorem}
\newtheorem{corollary}{Corollary}
\newtheorem{lemma}{Lemma}

\def\BibTeX{{\rm B\kern-.05em{\sc i\kern-.025em b}\kern-.08em
    T\kern-.1667em\lower.7ex\hbox{E}\kern-.125emX}}
\begin{document}

\title{Unequal Covariance Awareness for Fisher Discriminant Analysis and Its
Variants in Classification}

\author{\IEEEauthorblockN{Thu Nguyen $^{\star}$}
\IEEEauthorblockA{\textit{Department of Holistic Systems} \\
\textit{Simula Metropolitan}\\
Oslo, Norway \\
thu@simula.no}
\and
\IEEEauthorblockN{Quang M. Le $^{\star}$}
\IEEEauthorblockA{\textit{AISIA Research Lab}\\\textit{Department of Computer Science} \\
\textit{University Of Science}\\
\textit{Vietnam National University in Ho Chi Minh City}\\
Ho Chi Minh city, Vietnam \\
lequang.hvanhn@gmail.com}

\and
\IEEEauthorblockN{Son N.T. Tu}
\IEEEauthorblockA{\textit{Department of Mathematics} \\
\textit{University of Wisconsin-Madison}\\
Wisconsin, USA  \\
thaison@math.wisc.edu}
\and
\IEEEauthorblockN{Binh T. Nguyen}
\IEEEauthorblockA{\textit{AISIA Research Lab} \\
\textit{Department of Computer Science} \\
\textit{University Of Science}\\
\textit{Vietnam National University in Ho Chi Minh City}\\
Ho Chi Minh city, Vietnam \\
ngtbinh@hcmus.edu.vn}
}

\maketitle
\def\thefootnote{$\star$}\footnotetext{ denotes equal contribution}

\begin{abstract} 
Fisher Discriminant Analysis (FDA) is one of the essential tools for feature extraction and classification. In addition, it motivates the development of many improved techniques based on the FDA to adapt to different problems or data types. However, none of these approaches make use of the fact that the assumption of equal covariance matrices in FDA is usually not satisfied in practical situations. Therefore, we propose a novel classification rule for the FDA that accounts for this fact, mitigating the effect of unequal covariance matrices in the FDA. Furthermore, since we only modify the classification rule, the same can be applied to many FDA variants, improving these algorithms further. Theoretical analysis reveals that the new classification rule allows the implicit use of the class covariance matrices while increasing the number of parameters to be estimated by a small amount compared to going from FDA to Quadratic Discriminant Analysis. We illustrate our idea via experiments, which shows the superior performance of the modified algorithms based on our new classification rule compared to the original ones.
\end{abstract}

\begin{IEEEkeywords}
Fisher Discrimiant Analysis, Linear Discriminant Analysis, Quadratic Discriminant Analysis, classification
\end{IEEEkeywords}

\section{Introduction} \label{sec:introduction}


Fisher's Linear Discriminant Analysis (FDA) has long been an essential tool for feature extraction and classifications \cite{johnson2014applied}. Its core idea is to seek a series of projections that maximize the ratio of the between and within-class scatter matrices. During the computation of these matrices, it makes use of the label information. Thus, it is different from Principle Component Analysis, which does not account for the labels during dimension reduction.

Due to its effectiveness, there have been many efforts to adapt/improve the traditional FDA to different fields/situations. For example, Modified Fisher Discriminant Function \cite{mahmoudi2015detecting} is an FDA variant that uses weighted means that is more sensitive to the important instances and applied it to credit card fraud detection. In \cite{le2020adapted}, Le et al. proposed an adapted linear discriminant analysis with variable selection for the classification in high-dimension and applied the method to medical data. Some other works that tried to adapt FDA to different fields include  the works with application in health care \cite{le2020adapted, tougaccar2020application,ricciardi2020linear,banu2016predicting, prakash2021improved}, and facial recognition \cite{muslihah2020texture,najafi2021local,bhattacharyya2013face, rahulamathavan2012facial,satonkar2012face}. In addition, the nature of the data may also require adaption, which leads to even more modification of FDA. For example, to address the problem of outlier robustness in FDA, Oh et al. \cite{oh2013generalization} presented the $L_p$ norm linear discriminant analysis, which replaced $L_2$ norm in FDA with $L_p$ norm. 
Next, to address the small sample size problem of the FDA, various works have been done on sparse FDA \cite{witten2011penalized,qiao2009sparse,mai2012direct, chu2012sparse}. Another group of FDA variants is for FDA with imbalanced data \cite{chumachenko2021robust,chu2011characterization, jing2011class}. To deal with multimodal data, Sugiyama et al. \cite{sugiyama2006local} presented  Local Fisher Discriminant Analysis, and Kim et al. \cite{kim2010kernel} introduced kernel MFDA. 

In sum, it can be said that many FDA variants have been developed to deal with different types of problems/ data. 
Yet, none of these works has used the fact that the assumption of equal covariance matrices in the FDA is usually not valid in practical situations. Therefore, it motivates us to propose new classification rules for FDA and its variants. Moreover, as will be shown in the section ``Experiments'', incorporating that fact can significantly improve the modified versions compared to the corresponding original versions. 


The remaining of this work is organized as follows. First, in Section \ref{relatedWorks}, we review some related works on this topic. Next, Section \ref{fdareviews} reviews the traditional FDA and some related classical techniques. Then, we describe our framework and analyze its theoretical properties in Section \ref{uc-fda}. After that, Section \ref{experiments} demonstrate the power of our framework via experiments on various datasets using many FDA variants. Lastly, Section \ref{sec:conclusion} summarize the ideas and contribution of this works. 

\section{Related works}\label{relatedWorks}
\label{sec:related_work}

There have been many modifications to the original FDA. Many of them concentrate on modifying the within and between-class scatter matrix or defining a new weighted objective function \cite{duin2004linear,gu2015uncorrelated,sugiyama2007dimensionality, mahmoudi2015detecting}. Yet, many times, modifications are also made based on the target problems.

In order to address the problem of outlier robustness in FDA, Oh et al. \cite{oh2013generalization} suggested using the $L_p$ norm instead of $L_2$ norm and the steepest gradient to optimize the objective function. On the other hand, Ye et al. \cite{ye2018lp} presented 
$L_p$- and $L_s$-Norm Distance Based Robust Linear Discriminant Analysis. They used $L_p$ norm for the denominator and $L_s$ norm for the numerator of the objective function. Next, Yan and colleagues \cite{yan2010l} generalized Multiple Kernel Fisher Discriminant Analysis such that the kernel weights could be regularised with an $L_p$ norm for any $p\ge 1$. Some other related works can be Non-Sparse Multiple Kernel Fisher Discriminant Analysis \cite{yan2012non}, Fisher Discriminant Analysis with $L_1$-norm \cite{wang2013fisher}. Yet, the $L_p$ norm is harder to be optimized than the $L_2$ norm and may be computationally expensive, especially for big datasets.

Next, there have been various works to address the problem of small sample size compared to the number of features, well known as Sparse FDA \cite{qiao2009sparse,witten2011penalized,mai2012direct, chu2012sparse}. Penalized LDA \cite{witten2011penalized} is a general approach for penalizing the discriminant vectors in FDA using $L_1$ and Fused Lasso penalties in a way that leads to greater interpretability. As another example, Qiao et al. \cite{qiao2009sparse} developed a method for automatically incorporating variable selection in FDA. They applied regularization to obtain sparse linear discriminant vectors, where the discriminant vectors have only a small number of nonzero components. These methods have been successful in genetical datasets \cite{witten2011penalized,qiao2009sparse}.

Another group of FDA variants consists of the FDA variants for imbalanced data \cite{chumachenko2021robust,chu2011characterization, jing2011class}. Fast Subclass Discriminant Analysis and Subclass Discriminant Analysis \cite{chumachenko2021robust} allow one to put more attention on under-represented classes or
classes that are likely to be confused with each other.  \cite{chu2011characterization} focused on Uncorrelated Linear Discriminant Analysis for imbalanced data. Class-balanced Discrimination (CBD) and Orthogonal CBD (OCBD) \cite{jing2011class} are the two dimensional reduction techniques for imbalanced data.

For dealing with multimodal data, Sugiyama and the team \cite{sugiyama2006local} introduced  Local Fisher Discriminant Analysis for dimensionality reduction. Kim et al. \cite{kim2010kernel} proposed Kernel multimodal discriminant analysis and applied it to speaker verification, etc.

In addition, some other interesting modifications exist. In \cite{na2010linear}, Seng and colleagues recommended linear boundary discriminant analysis, which reflects the differences of non-boundary and boundary patterns. For big data,  Seng et al. \cite{seng2017big} proposed the SC-LDA algorithm replacing the full eigenvector decomposition of LDA with eigenvector decomposition on smaller sub-matrices. Then, they recombine the intermediate results to obtain the reconstruction. Finally, separability-oriented subclass discriminant analysis \cite{wan2017separability} divides every class into subclasses effectively to deal with the problem of a small number of features extracted when the number of classes is small. 

However, to our knowledge, there has not been any work that uses the fact that the assumption in FDA is usually not satisfied in a practical situation.

\begin{figure*}[!ht]
    \centering
    \includegraphics[scale = .72]{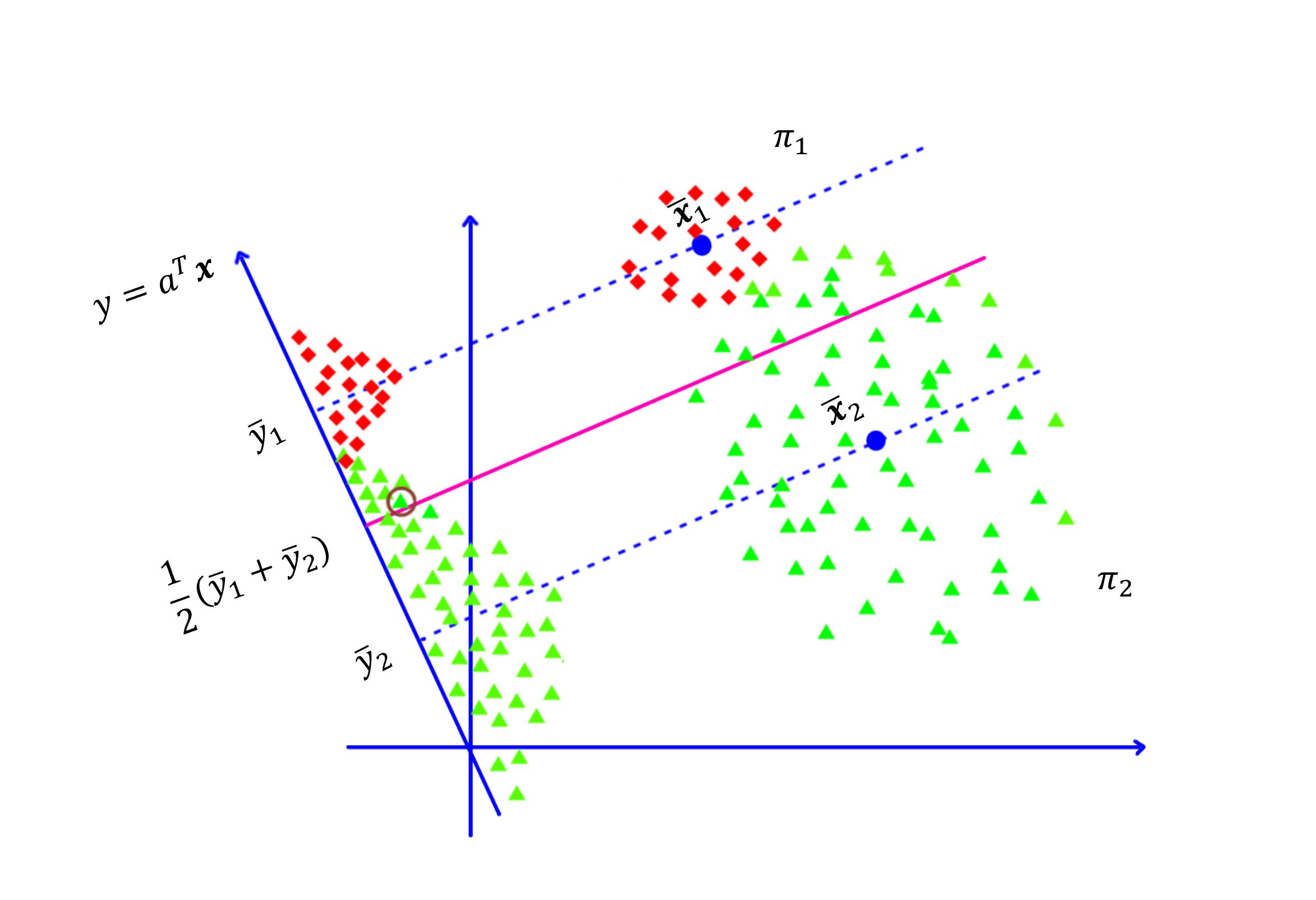}
    \caption{Motivating example for unequal convariance awareness.}
    \label{fig1}
\end{figure*}
\section{Preliminaries: Fisher Discriminant Analysis (FDA) and related methods}\label{fdareviews}
We denote by $\mathbf{a}^T$ the transpose of a vector $\mathbf{a}$. In this section, we briefly summarize the FDA and some related methods. We start by defining some notations.

Suppose that there are $C$ classes, where the $i^{th}$ class has $n_i$ observations, and $n=\sum_{i=1}^{C}n_i$ is the total number of samples. Denote by $\mathbf{x}_{ij}$ the $j^{th}$ observation from the $i^{th}$ class. 
Let 
\begin{equation}
	\bar{\mathbf{x} } = \frac{\sum_{i=1}^Cn_i\bar{\mathbf{x}}_i}{\sum_{i=1}^C n_i}
\end{equation}
be the overall mean and
\begin{equation}
	\bar{\mathbf{x} }_i = \frac{\sum_{j=1}^{n_i}{\mathbf{x}}_{ij}}{ n_i}
\end{equation}
be the mean of the $i^{th}$ class.

Next, let
\begin{align}
	\mathbf{B} &=\sum_{i=1}^Cn_i (\bar{\mathbf{x}}_i - \bar{\mathbf{x}})(\bar{\mathbf{x}}_i - \bar{\mathbf{x}})^T,\\
	\mathbf{W} &= \sum_{i=1}^C\sum_{j=1}^{n_i} (\mathbf{x}_{ij}-\bar{\mathbf{x}}_i )(\mathbf{x}_{ij} - \bar{\mathbf{x}}_{i})^T 
\end{align}
be the between-class and the within-class scatter matrix, respectively.

Now, we assume that $\mathbf{S}_{i}$ is the sample covariance matrix of the $i^{th}$ class, i.e.,
\begin{equation}\label{defSi}
	\mathbf{S}_{i} = \frac{1}{n_i-1}\sum_{j=1}^{n_i} (\mathbf{x}_{ij}-\bar{\mathbf{x}}_i )(\mathbf{x}_{ij} - \bar{\mathbf{x}}_{i})^T.
\end{equation}
\subsection{Fisher Linear Discriminant Analysis (FDA)}
FDA tries to find to projection $\mathbf{a}$ that maximizes the following Fisher criterion \cite{johnson2014applied}
\begin{equation} \label{def:trace}
	r = \frac{\mathbf{a}^T\mathbf{B}\mathbf{a}}{\mathbf{a}^T\mathbf{W}\mathbf{a}}.
\end{equation}
Let ${\lambda}_1 \ge {\lambda}_2 \ge ...\ge {\lambda}_s > 0$ be the $s\le \min(C-1,p)$ nonzero eigenvalues of $\mathbf{W}^{-1}\mathbf{B}$ and $\mathbf{v}_1, \mathbf{v}_2,...,\mathbf{v}_s$ be the corresponding normalized eigenvectors.


Suppose that we choose $r$ largest eigenvalues for classification. Then, we have $r$ corresponding projection space. Let $\mathbf{y}_j = \mathbf{v}_j^T\mathbf{x}$ be the projection of $\mathbf{x}$ onto the $j^{th}$ space, $j = 1,2,...,r$.  
Then, the sample mean of the $i^{th}$ class in the $j^{th}$ projection space is $  m_{ij} =  \mathbf{v}_j^T\bar{\mathbf{x}}_i$. 

The traditional FDA method allocates an observation $\mathbf{x}$ to $\pi_k$ if
\begin{equation}\label{eq36}
	 {\Sigma }_{j=1}^r  ({y}_j - {m}_{kj} )^2 \le  {\Sigma }_{j=1}^r ({y}_j - m_{ij})^2 \;\;\;\forall i \neq k.
\end{equation}
\subsection{Linear Discriminant Analysis (LDA)}
LDA is a commonly used classification technique that is usually mistaken with FDA. They both assume that the covariance matrices of all classes are equal. Nevertheless, unlike the FDA, which seeks a series of projections that maximize the ratio between-class and within-class scatter matrices, LDA assumes that the data from each class follows a multivariate Gaussian distribution and tries to minimize the total probability of misclassification  \cite{johnson2014applied}. The classification rule in LDA is to classify $\mathbf{x}$ to the $k^{th}$ class if 
\begin{equation}
	d_k(\mathbf{x} ) = \max \{d_1(\mathbf{x}),d_2(\mathbf{x} ),...,d_C(\mathbf{x}) \},
\end{equation}
where for $i = 1,2,...,C,$
\begin{equation}
	d_i(\mathbf{x}) = \bar{\mathbf{x}}_i^T \mathbf{S}_p^{-1} \mathbf{x} - \frac{1}{2}\bar{\mathbf{x}}_i^T \mathbf{S}_p^{-1} \bar{\mathbf{x}}_i +\log \frac{n_i}{n}.
\end{equation}
where $\mathbf{S}_p$ is the pooled covariance matrix, defined by
\begin{equation} \label{spdef}
    \mathbf{S}_p=\frac{\sum_{i=1}^C(n_i-1)\mathbf{S}_i}{\sum_{i=1}^Cn_i-C}
    = \frac{\mathbf{W}}{\sum_{i=1}^Cn_i-C}.
\end{equation}
Here, $\mathbf{S}_i$ is defined as in Equation (\ref{defSi}).
\subsection{Quadratic Discriminant Analysis (QDA)}
QDA also requires the data from each class to follow a multivariate Gaussian distribution as LDA. However, it does not assume that the covariance matrices are equal. The QDA classification rule is to classify $\mathbf{x}$ to the $k^{th}$ class if 
\begin{equation}
	d_k(\mathbf{x} ) = \max \{d_1(\mathbf{x}),d_2(\mathbf{x} ),...,d_C(\mathbf{x}) \},
\end{equation}
where for $i = 1,2,...,C,$
\begin{equation}
	d_i(\mathbf{x}) = -\frac{1}{2}\log|\mathbf{S }_i|-\frac{1}{2}(\mathbf{x}-\bar{\mathbf{x}}_i)^T\mathbf{S }_i ^{-1}(\mathbf{x}-\bar{\mathbf{x}}_i)+\log \frac{n_i}{n}.
\end{equation}

\section{Unequal covariance matrix awareness for FDA and its variants}\label{uc-fda}

This section will discuss the motivation and strategy for new classification rules.

\subsection{UC-FDA.}
The motivation of unequal covariance awareness can be explained via Figure \ref{fig1}. In this example, suppose that we have a binary classification task. Then, since there are only two classes, there exists only one projection. Let $\bar{y}_1$ denotes the mean of class $\pi_1$ in the projected space, $\bar{y}_2$ denotes the mean of class $\pi_2$ in the projected space, and
\begin{equation}
   \bar{y} = \frac{1}{2}(\bar{y}_1+\bar{y}_2).
\end{equation}
Then classification rule is to assign the observations on the left of $\bar{y}$ to class $\pi_1$ and the remaining to class $\pi_2$. Another equivalent classification strategy is to classify a sample $\mathbf{x}$ to $\pi_1$ if its projection $z$ has
\begin{equation}
    (z-\bar{y}_1)^2 \le (z-\bar{y}_2)^2.
\end{equation}
With such a classification rule, note that all the green sample on the left side of the violet line will be miss-classified into $\pi_1$. 

To be more specific, suppose that $\bar{y}_1=0, \bar{y}_2=3$ and the standard deviation of class $\pi_1, \pi_2$ in the projected space are $s_1 = 1, s_2 = 2$, respectively. Next, suppose that $\mathbf{x}$ is a $\pi_2$ sample whose projection in the projected space is $z=1.4$. Then, 
\begin{equation}
    (z-\bar{y}_1)^2 = 1.4^2,
\end{equation}
and
\begin{equation}
    (z-\bar{y}_2)^2 = 1.6^2,
\end{equation}
which resulted in $\mathbf{x}$ being missclassified into $\pi_1$. 

Meanwhile, if we take into account the variation of each class in the projected space then we can consider the distance between $z$ and $\bar{y}_1$ to be
\begin{equation}
    \left(\frac{z-\bar{y}_1}{s_1}\right)^2 = 1.96,
\end{equation}
and similarly, the distance between $z$ and $\bar{y}_2$:
\begin{equation}
    \left(\frac{z-\bar{y}_2}{s_2}\right)^2 = 0.64,
\end{equation}
which implies that $z$ is closer to $\bar{y}_2$ and $\mathbf{x}$ should be classified into $\pi_2$.


We formularize and extend the idea into the general case for a dataset with $C$ classes with all the above observations.
That is, we introduce unequal covariance awareness into the classification rule for FDA-based approaches. 

Here, we use the notations as in Section \ref{fdareviews}. Recall that the classification rule for FDA is given in Equation (\ref{eq36}). The \textit{unequal covariance awared} version of FDA, denoted as {UC-FDA}, is the same as the original FDA, except the classification rule is as follows.

Allocate the observation $\mathbf{x}$ to the $k^{th}$ population if
\begin{equation}\label{eq49}
	 {\Sigma }_{j=1}^r \frac{({y}_j - m_{kj})^2}{s_{kj}^2} \le  {\Sigma }_{j=1}^r \frac{({y}_j - m_{ij})^2}{s_{ij}^2} \forall i \neq k, 
\end{equation}
where $s_{ij}^2$ is the sample variance of the $i^{th}$ class in the $j^{th}$ projected space, i.e.,
\begin{equation}
s_{ij}^2 = \displaystyle\frac{1}{n_i - 1}\sum_{l=1}^{n_i}(y_{ilj} - m_{ij})^2.
\end{equation}
Here, $y_{ilj}$ is the projection of the  vector $\mathbf{x}_{il}$ (the $l^{th}$ sample from the $i^{th}$ class) onto the $j^{th}$ space. 

\textbf{\textit{Remarks.}} Since many modified versions of FDA such as Kernel Discriminant Analysis, Robust Fisher LDA \cite{kim2006robust}, {LDA-}$L_p$ \cite{oh2013generalization}, Incremental LDA \cite{pang2005incremental}, uncorrelated, weighted LDA \cite{liang2007uncorrelated}, Multiple Kernel Fisher Discriminant Analysis \cite{yan2010l} also apply the same classification rule as in FDA, this modification scheme could also be applied to these methods. 


\subsection{Theoretical analysis}\label{theoraticalAna}
\begin{table*}[ht]
	\caption{Descriptions of data sets used in the experiments}
	\vspace*{3mm}
	\centering
	\label{tab1}
		\begin{tabular}{|c|c|c|c|}
			\hline
			\textbf{Datasets}  &  \textbf{\#classes}  & \textbf{ \#features}  & \textbf{\#samples} \\\hline
			Heart & 2 & 44& 267\\\hline
			Car &  4 & 6 & 1728 \\\hline
			Balance & 3 & 4 & 625\\\hline
			\begin{tabular}{@{}c@{}}Breast \\ tissue\end{tabular}& 6 & 9 & 106\\\hline
			{Digits} & 10 & $64\;(54^*)$ & 1797\\\hline
			{Seeds} & 3 & 7 & 210\\ \hline
			{Wine}& 3 & 13 & 178 \\        \hline
			{Iris}& 3 &  4 &  150  \\  \hline
			CNAE-9 & 2 & 60 & 208  \\\hline
			Glass & 6 & 9 & 214 \\\hline
	\end{tabular}
\end{table*}
In this section, we analyze our methodology via the traditional ${L} _2$-norm FDA and its covariance aware version.

As simple as the approach may sound, our framework possesses some nice properties.

\subsubsection{Implicit use of the covariance matrices and analysis of number of parameters}\label{implicit}
\begin{theorem}
Let $\mathbf{v}_j$ is the $j^{th}$ eigenvector of $\mathbf{W}^{-1}\mathbf{B}$. Then $ s_{ij}^2$, the sample variance of the projections of the  $i^{th}$ class observations into the $j^{th}$ space, satisfies
\begin{equation}
    s_{ij}^2=\mathbf{v}_j^T\mathbf{S }_{i}\mathbf{v}_j.
\end{equation}
\end{theorem}

\begin{IEEEproof}
By definition, the sample variance of the $i^{th}$ class in the $j^{th}$ projected space is
\begin{equation}
s_{ij}^2 = \displaystyle\frac{1}{n_i - 1}\sum_{l=1}^{n_i}(y_{ilj} - m_{ij})^2,
\end{equation}
where $y_{ilj}$ is the projection of the  vector $\mathbf{x}_{il}$ (the $l^{th}$ sample from the $i^{th}$ class) onto the $j^{th}$ space, $m_{ij}$ is the sample mean of the $i^{th}$ class in the aforementioned space, and $n_i$ is the sample size of the $i^{th}$ class. 

Also, we have 
\begin{equation}
    y_{ilj} = \mathbf{v}_{j}^{T}\mathbf{x}_{il}
\end{equation}
and 
\begin{equation}
    m_{ij} = \mathbf{v}_{j}^{T}\mathbf{\bar{x}}_{i}.
\end{equation}

Thus,
\begin{equation}
\begin{split}
    s_{ij}^2 &=\displaystyle\frac{1}{n_i-1}\sum_{l=1}^{n_i}(\mathbf{v}_j^T\mathbf{x}_{il} - \mathbf{v}_j^T\bar{\mathbf{x}}_i)^2\\
    &= \frac{1}{n_i-1}\sum_{l=1}^{n_i}\mathbf{v}_j^T(\mathbf{x}_{il} -\bar{\mathbf{x}}_i)
	(\mathbf{x}_{il} -\bar{\mathbf{x}}_i)^T\mathbf{v}_j\\
	&=\mathbf{v}_j^T \left\{\frac{1}{n_i-1}\sum_{l=1}^{n_i}(\mathbf{x}_{il} -\bar{\mathbf{x}}_i)
	(\mathbf{x}_{il} -\bar{\mathbf{x}}_i)^T\right\}\mathbf{v}_j.\\
\end{split}
\end{equation}
Therefore, using Equation (\ref{defSi}), we have
\begin{equation}
   s_{ij}^2 = \mathbf{v}_j^T\mathbf{S}_{i}\mathbf{v}_j,
\end{equation}
which ends our proof.
\end{IEEEproof}

From this theorem, we have the following corollary 
\begin{corollary}
Let $\mathbf{x}$ be an observation and $y_{j}=\mathbf{v}^T_j \mathbf{x}$ where $\mathbf{v}_j$ is the $j^{th}$ eigenvector of $\mathbf{W}^{-1}\mathbf{B }$. 
Suppose that we select only the first $r$ non-zero eigenvectors of $\mathbf{W}^{-1}\mathbf{B }:$ $\mathbf{v}_1,..., \mathbf{v}_r$ for classification.  Then
\begin{equation}\label{eq11}
	{\Sigma }_{j=1}^r \frac{({y}_j - m_{ij})^2}{s_{ij}^2} =  {\Sigma }_{j=1}^r \frac{\left[\mathbf{v} _{j}^T\left(\mathbf{x} -\bar{\mathbf{x}}_i \right)\right]^{2}}{\mathbf{v}_j^T\mathbf{S }_{i}\mathbf{v}_j}.
\end{equation}

\end{corollary}



From the above theorem and corollary, we see that even though we don't use the estimates of $\mathbf{\Sigma}_i$ as in QDA, we implicitly use them via classification rule. This is a very nice property of our framework because this allows making use of $\mathbf{\Sigma}_i$ without increasing the number of parameters to be estimated by a significant as going from FDA to QDA. 

Specifically, for a classification task with $G$ classes, if $r$ eigenvalues are selected for classification, our methods have $r\times G$ more parameters to estimate than the FDA. Yet, this increment is minuscule compared to switching from FDA to QDA ($(G-1)\times p^2+p$ more parameters), where we have to estimate the covariance matrix for each class.  
Therefore, it is a cheap and worthy trade-off compared to going from FDA to QDA.
\subsubsection{Relation to QDA, FDA, LDA and Mahalanobis distance}
Recall that QDA classifies $\mathbf{x}$ to the $k^{th}$ class if $d_k(\mathbf{x})$ is the smallest among
\begin{equation}\label{52qda}
	d_i(\mathbf{x}) = -\frac{1}{2}\log|\mathbf{S }_i|-\frac{1}{2}(\mathbf{x}-\bar{\mathbf{x}}_i)^T\mathbf{S }_i ^{-1}(\mathbf{x}-\bar{\mathbf{x}}_i)+\log \frac{n_i}{n}, 
\end{equation}
where $i = 1,...,C.$

Aslo, LDA classifies $\mathbf{x}$ to the $k^{th}$ class if $d_k(\mathbf{x})$ is the smallest among
\begin{equation}\label{52nlda}
	d_i(\mathbf{x}) = \bar{\mathbf{x}}_i^T \mathbf{S}_p^{-1} \mathbf{x} - \frac{1}{2}\bar{\mathbf{x}}_i^T \mathbf{S}_p^{-1} \bar{\mathbf{x}}_i +\log \frac{n_i}{n},
\end{equation}
for $i = 1,...,C.$

In addition, the FDA assigns a sample $\mathbf{x}$ to the $k^{th}$ if $d_k(\mathbf{x})$ is the smallest among
\begin{equation}\label{eq523}
   d_i(\mathbf{x}) = \sum_{j=1}^p [\mathbf{v}_j^T(\mathbf{x}-\bar{\mathbf{x}}_i)]^2 = (\mathbf{x}-\bar{\mathbf{x}}_i)^T\mathbf{S}_{p}^{-1}(\mathbf{x}-\bar{\mathbf{x}}_i), 
\end{equation}
for $i = 1,...,C.$

The proof of the relation in Equation (\ref{eq523}) could be found in \cite{johnson2014applied}.


Hence, we can see that QDA, FDA can be considered as relying on Mahalanobis distance.
 On the other hand, our classification rule has the following property,
\begin{theorem}\label{thr42}
Suppose that we select only the first $r$ non-zero eigenvectors of $\mathbf{W}^{-1}\mathbf{B }:$ $\mathbf{v}_1,..., \mathbf{v}_r$ for the classification. Then,
\begin{align}\label{eq10}
	{\Sigma }_{j=1}^r \frac{({y}_j - m_{ij})^2}{s_{ij}^2} &=  {\Sigma }_{j=1}^r \frac{\left[\mathbf{v} _{j}^T\left(\mathbf{x} -\bar{\mathbf{x}}_i \right)\right]^{2}}{\mathbf{v}_j^T\mathbf{S }_{i}\mathbf{v}_j}\\
	&\le r (\mathbf{x}-\bar{\mathbf{x}}_i)^T\mathbf{S}^{-1}_i (\mathbf{x}-\bar{\mathbf{x}}_i)
\end{align}
\end{theorem}
The proof follows directly from the following result \cite{johnson2014applied}:
\begin{lemma}{(Extended Cauchy-Schwarz inequality)}
Let $\mathbf{b}, \mathbf{d}\in \mathbb{R}^p$ be any two vectors, and let $\mathbf{B} \in \mathbb{R}^{p\times p}$ be a positive definite matrix. Then
\begin{equation}
\left(\mathbf{b}^T \mathbf{d}\right)^{2} \leq\left(\mathbf{b}^T \mathbf{B} \mathbf{b}\right)\left(\mathbf{d}^T \mathbf{B}^{-1} \mathbf{d}\right)
\end{equation}
with the equality if and only if $\mathbf{b}=c \mathbf{B}^{-1} \mathbf{d}$ or $(\mathbf{d}=c \mathbf{B} \mathbf{b})$ for some constant $c$.
\end{lemma}

Moreover, from Equations (\ref{52qda}), (\ref{52nlda}), and (\ref{eq523}), we can see that the classification rules for LDA, FDA, QDA all involves the matrix inversion of the sample covariance matrices or pooled covariance matrix. Meanwhile, from Equation (\ref{eq10}), we see that FDA does not have such a requirement. In addition, $s_{ij}^2$ can be estimated empirically. Therefore, FDA has the advantage of no large matrix inversion for large datasets.

\begin{table*}[ht]
\centering
\caption{The 5-fold cross-validation results using FDA, UC-FDA, QDA. The bold denotes the one that best performs among FDA and UC-FDA. Note that one value in the QDA column is not available, denoted by NA. This is due to the division by zero error encountered by Sklearn \cite{sklearn_api}.}\label{tab:fda}
\vspace*{3mm}
\begin{tabular}{|c|c|c|>{\columncolor{lightgray}} c|}
\hline
\textbf{Datasets} & \textbf{FDA} & \textbf{UC-FDA} & \textbf{QDA} \\ \hline
Heart & 0.354 & \textbf{{0.296}} & \textit{\textbf{0.208}} \\ \hline
Car & 0.509 & \textbf{0.380} &  NA   \\ \hline
Balance & 0.256 & \textbf{0.084} & \textit{\textbf{0.084}}\\\hline
\begin{tabular}{@{}c@{}}Breast \\ tissue\end{tabular} & 0.388 & \textbf{0.356} & 0.388\\\hline
Digits & 0.053 & \textbf{0.051} & 0.122\\\hline
Seeds & 0.096 & \textbf{0.038} & 0.059\\\hline
Wine & 0.345 & \textbf{0.271} & \textit{\textbf{0.017}} \\\hline
Iris & \textbf{0.027} & \textbf{0.027} & \textbf{\textit{0.014}}\\\hline
CNAE-9 & 0.269 & \textbf{0.244} & 0.355\\\hline
Glass & 0.466 & \textbf{0.426} & NA\\\hline
\end{tabular}
\end{table*}

\section{Experiments}\label{experiments}
\subsection{Methods under comparision}
Recall that we denoted FDA as the traditional Fisher Discriminant Analysis, and UC-FDA is its unequal covariance aware version. In addition, let \textbf{SDA} be the Fisher Discriminant Analysis with covariance shrinkage \cite{chen2010shrinkage}, we denote by \textbf{UC-SDA} its unequal covariance aware version. Moreover, let \textbf{LDA-$L_p$} be the Generalization of linear discriminant analysis using Lp-norm \cite{oh2013generalization}, we denote by \textbf{UC-LDA-$L_p$} its unequal covariance aware version. We will compare the performance of these algorithms. 

Note that FDA is already described in Section \ref{fdareviews}. Therefore, in the followings, we give some short descriptions about SDA and {UC-LDA-$L_p$},
\begin{itemize}
    \item \textbf{SDA} is a variant of Fisher Discriminant Analysis where the sample covariance matrices are replaced with the corresponding covariance shrinkage estimate \cite{chen2010shrinkage}, which leads to the following modification of the within-class scatter matrix in its unequal covariance aware (UC-SDA) version
    \begin{equation}
        \mathbf{W} = \sum_{i=1}^Cn_i \mathbf{S}_{iShrink}
    \end{equation}
    where $\mathbf{S}_{iShrink}$ is the covariance shrinkage estimate of the $i^{th}$ class, $n_i$ is the number of samples that belong to the $i^{th}$ class, and $C$ is the number of classes. 
    
    \item \textbf{LDA-$L_p$} \cite{oh2013generalization} is a generalization of FDA that uses an  $L_p$-norm instead of $L_2$ norm in both the numerator and denominator of the objective function. Using our notations, the objective function to be maximized can be written as
    \begin{equation}
        F(\mathbf{w}) = \frac{\sum_{i=1}^Cn_i|\bar{\mathbf{x}}_i-\bar{\mathbf{x}}|^p}{\sum_{i=1}^C\sum_{j=1}^{n_i}|\mathbf{w}^T(\mathbf{x}_{ij}-\bar{\mathbf{x}}_i)|^p},
    \end{equation}
    where $\mathbf{w}$ is projection vector. LDA-$L_p$ constraint that $||\mathbf{w}||_2=1$ and uses steepest gradient as the optimization tool. 
\end{itemize}
\subsection{Datasets and Implementation}

\begin{table*}[]
\centering
\caption{The 5-fold cross-validation results using SDA and UC-SDA. The bold indicates the best performance.}\label{tab:sda}
\vspace*{2mm}
\begin{tabular}{|>{\centering\arraybackslash}p{2cm}|c|c|}
\hline
\textbf{Datasets} & \textbf{SDA} & \textbf{UC-SDA} \\ \hline
Heart & 0.359 & \textbf{0.262}  \\ \hline
Car & 0.501 & \textbf{0.386} \\ \hline
Balance &  0.258 & \textbf{0.084} \\\hline
Breast  tissue & 0.369 & \textbf{0.294}\\\hline
Digits & \textbf{0.047} & 0.049 \\\hline
Seeds & 0.092 & \textbf{0.039} \\\hline
Wine & {0.302} & \textbf{0.231} \\\hline
Iris & 0.037 & \textbf{0.029} \\\hline
CNAE-9 & 0.419 & \textbf{0.243} \\\hline
Glass & 0.476 & \textbf{0.453} \\\hline
\end{tabular}
\end{table*}
Table \ref{tab1} shows a summary of all data sets used in the experiment, all of which comes from the Machine Learning Database Repository at the University of California, Irvine \cite{Dua:2019}.  For Digits, we delete ten columns where the number of nonzero values is less than 10 to avoid the issue with covariance inversion.

For each data set, we transform each feature by scaling and translating each feature individually such that it is between zero and one. 
For LDA-$L_p$ and UC-LDA-$L_p$, due to the computational cost of $L_p$ norm optimization, the number of projections used is 1, and the $\epsilon$ used for convergence check is $10^{-5}$, and the $L_p$ norm used has $p=1.5$. 

To examine whether the datasets satisfy the equal covariance matrix assumption in FDA, we also provide the results of the Box-M Test using Pingouin package \cite{Vallat2018}. Note that if the covariance matrices of all classes are equal, then the variance of the $i^{th}$ features of all classes are equal. Therefore, if $\log 0$ is encountered in Box's M test, we use the Levene test to check if there are features of which all classes' variances are not equal. At a significant level $\alpha = 0.05$, the hypotheses of equal covariance matrices or variance are rejected for all the datasets in the experiments.

The experiments are run directly on \textit{Google Colaboratory}\footnote{\url{https://colab.research.google.com/}}, and we will release the codes are available at \url{https://github.com/thunguyen177/UC-FDA}. 

\subsection{Evaluation Metrics}
For evaluation, we use K-fold cross-validation with $K = 5$. Here, the error rate is defined as the ratio between the number of miss-classification items and the total number of samples, i.e.,
\begin{equation}
    \text{error rate} = \frac{\#\; \text{missclassification}}{\# \;\text{samples}}.
\end{equation}

\subsection{Results and Analysis}
\begin{table*}[]
\centering
\caption{5-fold cross-validation results using LDA-$L_p$ and UC-LDA-$L_p$. The bold indicates the best performance}\label{tab:ldalp}
\vspace*{2mm}
\begin{tabular}{|>{\centering\arraybackslash}p{2cm}|c|c|}
\hline
\textbf{Datasets}& \textbf{LDA-$L_p$} & \textbf{UC-LDA-$L_p$} \\ \hline
Heart  & 0.337 & \textbf{0.272} \\ \hline
Car  & 0.625 & \textbf{0.270 }\\ \hline
Balance  & 0.313 & \textbf{0.238}\\\hline
Breast  tissue& 0.540 &\textbf{ 0.519}\\\hline
Digits & \textbf{0.604} & 0.630\\\hline
Seeds  & 0.206 & \textbf{0.205}\\\hline
Wine  & 0.060 & \textbf{0.053}\\\hline
Iris & \textbf{0.026} & 0.041\\\hline
CNAE-9 & 0.302 & \textbf{0.207}\\\hline
Glass & 0.544 & \textbf{0.432}\\\hline
\end{tabular}
\end{table*}
The results are as shown in Table \ref{tab:fda}, Table \ref{tab:sda}, and Table \ref{tab:ldalp}.

From Table \ref{tab:fda}, we see that the unequal covariance aware version of FDA can improve the original FDA by a significant amount. For example, for the Balance data set, UC-FDA has an error rate of $0.084$ compared to FDA at $0.256$. For the sake of exploring, we also report the result of QDA in the table. Note that the best performer among FDA and UC-FDA is marked in bold, and if QDA is the best performer, it is marked in bold italic. With that, one can see that UC-FDA often outperforms both FDA and QDA.

Another interesting point in Table \ref{tab:fda} is that UC-FDA performs even better than QDA (7.1\% better) for the Digits dataset, even though this dataset has 1797 samples and 64 features. That may be because the number of parameters to be estimated is much smaller than QDA, and the computation of UC-FDA does not involve inversing any large matrix for big datasets. This is consistent with what was discussed in the theoretical section.

Next, from Tables \ref{tab:sda} and \ref{tab:ldalp}, we can see that many times, the unequal covariance aware version of LDA-$L_p$ and SDA outperform the corresponding original version by a significant margin. For example, in the Heart data set, UC-SDA has an error rate of 0.262, which is a 9.7\% of error rate reduction for the original SDA, whose error rate is 0.359. With the same data set, the error rate of UC-LDA-$L_p$ is 0.272, which is a 6.5\% of error rate reduction for the original LDA-$L_p$, whose error rate is 0.337.

However, from these tables, we can see that the unequal covariance aware versions do not always outperform the corresponding original versions. This could depend on the FDA variant used or due to the increment in the number of parameters to be estimated leads to more computation error, while the variances in the projected spaces are not too different. Nevertheless, even in cases where the unequal covariance aware versions do not outperform the corresponding original versions, one can see that there is not much degradation in performance. As an example, in Table \ref{tab:sda}, for Digits, UC-FDA only increases the error rate by 0.2\%, and for Iris,  UC-FDA only increases the error rate by 0.8\%.



\section{Conclusion and Future Works}\label{sec:conclusion}
In this paper, we have discussed a simple technique to improve many variants of Fisher Discriminant Analysis. In addition, we showed that the new classification rule allows the implicit use of the class covariance matrices while increasing the number of parameters to be estimated by only a little compared to going from FDA to Quadratic Discriminant Analysis. We also illustrate via experiments the significant error reduction margins that our novel classification rule can achieve compared to the original FDA variants. 

However, it is worth noting that the proposed framework does increase the number of parameters. Therefore, when the sample size is too small and/or the variances in the projected spaces are only slightly different, the classical approaches may outperform the UC methods. Though, even in those cases, the performance of classical techniques may only be marginally better than UC methods, as illustrated in the experiments.  

Another essential point to draw out from the paper is that when the assumption of a model is not satisfied in practice, it is worth exploring how to use that fact to improve the technique. Therefore, it would be interesting to explore how to extend this idea to different methods in the future. For example, in Normal Linear Discriminant Analysis, the covariance matrices are also assumed to be equal, which is usually not true in practical situations. So, it is worth examining how to incorporate that knowledge to boost performance even further. 

\section*{Acknowledgment}

We want to thank the University of Science, Vietnam National University in Ho Chi Minh City,  AISIA Research Lab in Vietnam, and SimulaMet for supporting us throughout this paper. 
The fourth author is supported by Vietnam National University Ho Chi Minh City (VNU-HCM) under grant number C2021-18-03.

\bibliographystyle{IEEEtran}
\bibliography{IEEEabrv,FDA}

\end{document}